\documentclass[letterpaper, 10 pt, conference]{ieeeconf}  % Comment this line out if you need a4paper

\IEEEoverridecommandlockouts                              % This command is only needed if 
\title{\LARGE \bf
FarSee-Net: Real-Time Semantic Segmentation by Efficient Multi-scale Context Aggregation and Feature Space Super-resolution 
}

\author{Zhanpeng Zhang$^{1}$, Kaipeng Zhang$^{2*}$\\
$^1${SenseTime Group Limited} $^2$The University of Tokyo\\
\thanks{*Corresponding author}%
}
\usepackage{graphicx}
\usepackage{amsfonts,amssymb}
\usepackage{url}
\begin{document}

\newcommand{\etal} {\textit{et al.}}
\newcommand{\ie} {\textit{i.e.}}
\newcommand{\eg} {\textit{e.g.}}
\maketitle
\thispagestyle{empty}
\pagestyle{empty}

%%%%%%%%%%%%%%%%%%%%%%%%%%%%%%%%%%%%%%%%%%%%%%%%%%%%%%%%%%%%%%%%%%%%%%%%%%%%%%%%
\begin{abstract}
Real-time semantic segmentation is desirable in many robotic applications with limited computation resources. One challenge of semantic segmentation is to deal with the object scale variations and leverage the context. How to perform multi-scale context aggregation within limited computation budget is important. In this paper, firstly, we introduce a novel and efficient module called Cascaded Factorized Atrous Spatial Pyramid Pooling (CF-ASPP). It is a lightweight cascaded structure for Convolutional Neural Networks (CNNs) to efficiently leverage context information. On the other hand, for runtime efficiency, state-of-the-art methods will quickly decrease the spatial size of the inputs or feature maps in the early network stages. The final high-resolution result is usually obtained by non-parametric up-sampling operation (e.g. bilinear interpolation). Differently, we rethink this pipeline and treat it as a super-resolution process. We use optimized super-resolution operation in the up-sampling step and improve the accuracy, especially in sub-sampled input image scenario for real-time applications. By fusing the above two improvements, our methods provide better latency-accuracy trade-off than the other state-of-the-art methods. In particular, we achieve 68.4\% mIoU at 84 fps on the Cityscapes test set with a single Nivida Titan X (Maxwell) GPU card. The proposed module can be plugged into any feature extraction CNN and benefits from the CNN structure development.
\end{abstract}

%%%%%%%%%%%%%%%%%%%%%%%%%%%%%%%%%%%%%%%%%%%%%%%%%%%%%%%%%%%%%%%%%%%%%%%%%%%%%%%%
\section{INTRODUCTION}
Semantic segmentation refers to the problem of estimating the class label for all pixels given the input image. It is a fundamental task for many computer vision applications. Thanks to the remarkable development of deep convolutional networks~\cite{resnet,ren2015faster,maskrcnn,krizhevsky2012imagenet,573696ce6e3b12023e5ce95a}, there is substantial progress for this task. However, many algorithms require sophisticated models that come with large memory and computational cost. This is challenging for robotics applications where real-time performance is required and the computation resource is usually limited. A better latency-accuracy trade-off is worth investigation. 

\begin{figure}[t]
	\centering
	\includegraphics[width=0.8\linewidth]{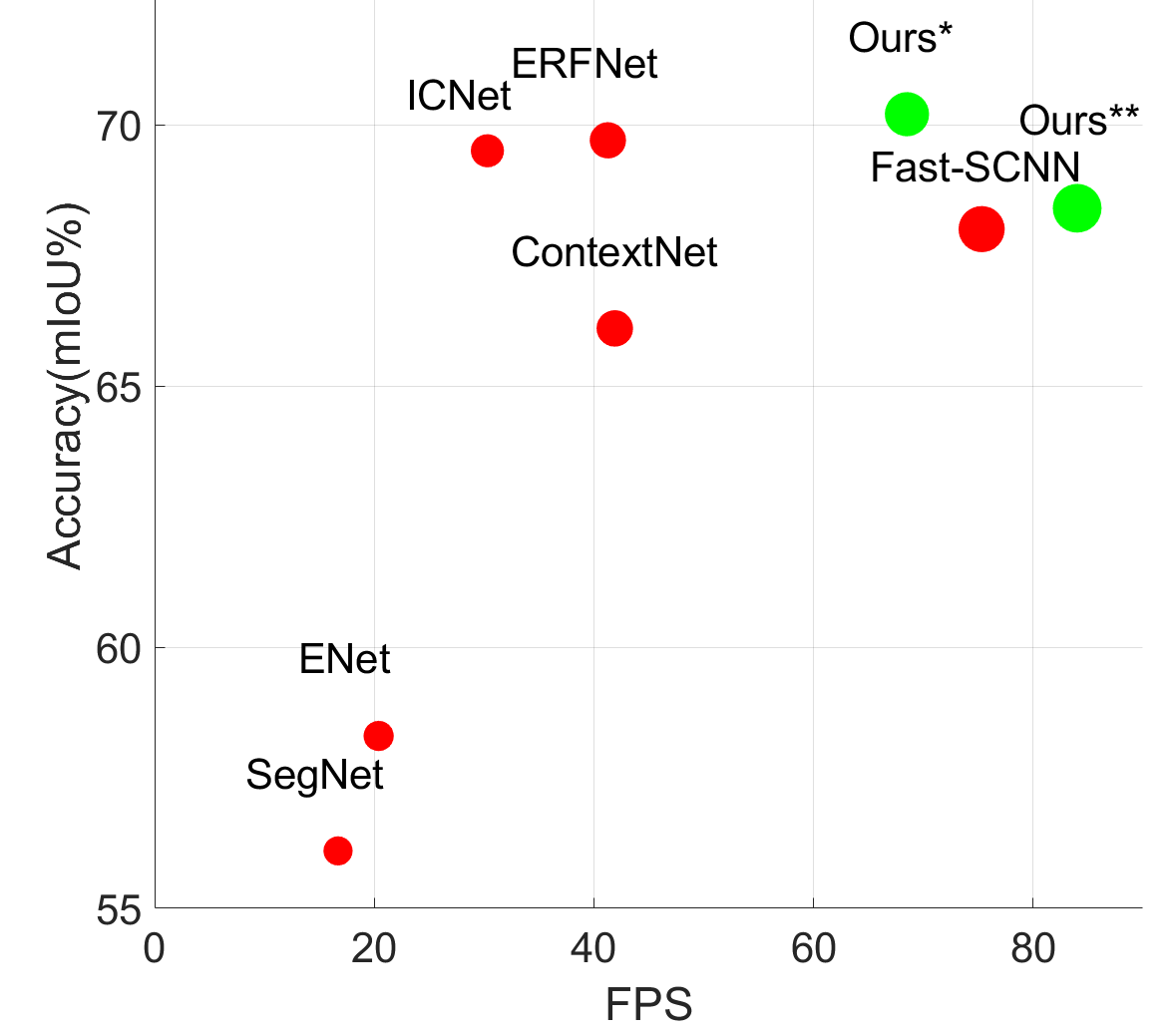}
	\caption{Illustration of the inference speed (fps) and accuracy (mIoU) on the Cityscapes test set. Bigger dot means faster inference process. The inference speed is evaluated on a single Nvidia Titan X (Maxwell) GPU, for the current real-time semantic segmentation methods, including SegNet~\cite{segnet}, ICNet~\cite{599c7972601a182cd263ec00}, ENet~\cite{enet}, ERFNet~\cite{erfnet} and Fast-SCNN~\cite{fastscnn}. `Ours*' and `Ours**' denotes our approach with different input image resolutions (\ie., $512\times 1024$ and $512\times 768$, respectively). The results of other methods come from the related literature.}
	\label{fig:result_fig}
\end{figure}

Current deep learning based semantic segmentation algorithm usually contains a front-end network and a back-end network. A backbone network pre-trained with large-scale image classification task is usually employed as the front-end for feature extraction. The back-end network, on the other hand, usually contains a module for multi-scale context aggregation (e.g., ASPP~\cite{deeplab,deeplabv3p}, PPM~\cite{pspnet}, RefineNet~\cite{refinenet}), as well as some sequential convolution layers to generate the final dense class probability map. Since the front-end network needs to extract high level semantic information, it usually requires deep networks with large numbers of parameters. To accelerate the algorithm, currently real-time segmentation methods employ two and multiple branch architecture~\cite{hrnet,599c7972601a182cd263ec00,bisenet} for the front-end. In particular, branch with deeper network is fed with a lower resolution version of the image while branch with shallower network deals with the higher resolution version. On the other hand, some approaches achieve real-time performance by designing thin network structure~\cite{fastscnn,erfnet}. But accuracy of the segmentation result usually degrades substantially with real-time acceleration techniques. Better trade-off between accuracy and speed stills requires investigation. Fig.~\ref{fig:result_fig} shows an overview about the speed and accuracy of some current real-time semantic segmentation methods.

Since the images may contain a same class in different scales, how to leverage the context information is important for thin and small objects. Multi-scale context aggregation with deep network plays an important role for semantic segmentation. In fact, the back-end network usually needs to accomplish this. Current context aggregation modules, such as Atrous Spatial Pyramid Pooling (ASPP)~\cite{deeplab,deeplabv3p}, have shown the effectiveness and rank among the state-of-the-art methods. However, these modules usually work in deep layers of the network where the number of feature map channel is large. In this case, even a convolutional layer with kernel size 3 consumes substantial computation. In this work, we designed a factorized ASPP module for multi-scale context fusion. Moreover, since this module is light-weight, we can repeatedly employ it for stronger context fusion without obvious increase of computation. Here we named this module cascaded factorized ASPP (CF-ASPP). 

\begin{figure}[t]
	\centering
	\includegraphics[width=\linewidth]{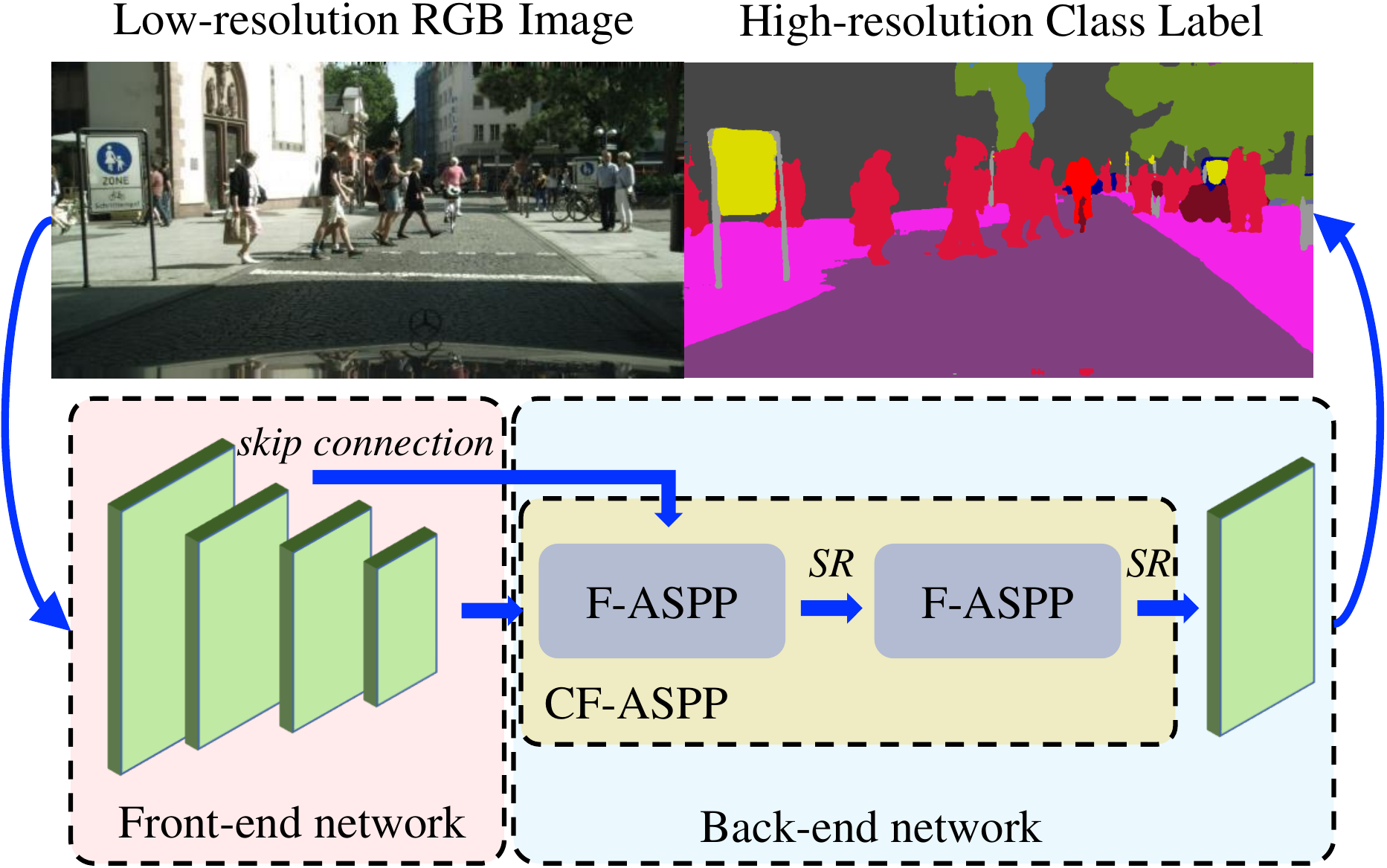}
	\caption{We divide the semantic segmentation network into front-end and back-end. For the front-end, we can use a generic image representation backbone network. For the back-end, we design a cascaded factorized ASPP (CF-ASPP) module that can perform multi-scale context aggregation efficiently, and combine it with feature space super-resolution (SR) learning, so as to boost the process of generating high-resolution output label map from low-resolution input. The proposed back-end network is easy to implement and can be plugged into other front-end networks.}
	\label{fig:idea}
\end{figure}

Another issue of the back-end network for semantic segmentation is that the spatial size of the feature maps decreases substantially after the front-end. In addition, many approaches improve the speed by using lower-resolution images as input directly. This makes it more challenging for the back-end network. Many current back-end networks simply perform up-sample by parameter-free operations such as bilinear interpolation, on the feature map to obtain the result with original resolution. In this case, it is hard to recover the details for the final segmentation result. Mazzini~\cite{gun} propose Guided Up-sampling Network (GUN) to better recover from the low-resolution feature map. In particular, the network learns to predict a high-resolution guidance offset table of offsets vectors that steer sampling towards the correct semantic class. In this work, we solve this problem from another perspective. In the training process, we use lower-resolution input image (\eg, $512\times 1024$ for Cityscapes dataset instead of $1024\times 2048$) but keeps the high-resolution ground truth for supervision. In fact, this is the problem of recovering high-resolution output from low-resolution input. This problem has been widely studied under the topic of super-resolution~\cite{srcnn}. We can leverage the development of super-resolution to solve this problem. This process can be performed by current highly optimized operators, which we fuse with the proposed CF-ASPP.

To this end, in this work, we propose a new back-end network for semantic segmentation. In particular, we propose the CF-ASPP module to efficiently perform multi-scale context aggregation and employ a feature map super-resolution step that can better recover high-resolution result from low-resolution input. Our whole pipeline is shown in Fig.~\ref{fig:idea}. To summarize, our contributions are three-fold: 
\begin{enumerate}
	\item We propose a cascaded factorized ASPP (CF-ASPP) module that is faster and provides more accurate result than the original ASPP~\cite{deeplabv3p,deeplab}.
	\item We treat the problem of recovering high-resolution segmentation result from low-resolution input as a super-resolution process. The experimental result shows that given lower resolution input image, the performance degrade of our method is lower than other methods. This helps to accelerate the algorithm while keeping reasonable accuracy.
	\item We provide a new back-end network for semantic segmentation. The proposed network provides better latency-accuracy trade-off than current state-of-the-art real-time semantic segmentation methods~\cite{fastscnn,599c7972601a182cd263ec00,bisenet}. In addition, the proposed back-end network is easy to implement and can be directly combined with other existing feature extraction network. That means our approach can benefit from the advance of common feature extraction network which is another important computer vision topic~\cite{mobinenetv2,shufflenet,espnetv2}. 
\end{enumerate}

\section{RELATED WORK}
\subsection{Quality Driven Semantic Segmentation}
Since the introduction of FCN~\cite{fcn}, there are extensive research works on deep learning for semantic segmentation. How to model the spatial relationship between pixels or leverage the context for inference is the main concern of current methods. For example, the skip-connection~\cite{resnet} is widely employed to fuse the high level semantic feature and low level spatial cues; ASPP~\cite{deeplab,deeplabv3,deeplabv3p} and PPM~\cite{pspnet} is the component applied on the extracted feature from front-end to fuse context of multiple scales; CRF~\cite{chen2015semantic} and MRF~\cite{liu2015semantic} is used to model the spatial relationship between pixels or regions. Recently, HRNet~\cite{hrnet} is proposed for semantic segmentation. The whole network maintains multiple branches with different resolutions of the image. The representation of different resolutions is densely fused. These methods achieve high quality results at the cost of heavy computation.

\subsection{Real-Time Semantic Segmentation}
Recently, real time semantic segmentation attracts more and more researchers. Research works among this line aims to improve the model inference time while keeping decent accuracy. ENet~\cite{enet} is one of the pioneers in this line. The authors design a light-weight network structure and the input image is heavily down-sampled at the early stage of the network to reduce processing time. ERFNet~\cite{erfnet} takes another approach, where the network contains multiple factorized convolution blocks. In particular, the combinations of $n\times1$ and $1\times n$ convolution is used to reduce the computation of original $n\times n$ convolution. Similar structure is also used in~\cite{lednet}. Two-branch and multi-branch network design is also proposed. ICNet~\cite{599c7972601a182cd263ec00}, ContextNet~\cite{contextnet}, BiSeNet~\cite{bisenet} and Fast-SCNN~\cite{fastscnn} learn global image context with lower resolution input with a deeper network branch, which is then combined with the feature from a shallower branch that describes the boundary information using higher resolution input, or feature map obtained from the early stage of the deeper branch. Reusing feature is another direction to reduce the computation cost. Most recently, Li~\etal~\cite{dfanet} design a network that contains multiple sub-modules. Features from early module is reused for the following module. Different from these approaches, we treat the whole segmentation network as a combination of front-end and back-end, and our work does not need specific backbone/feature-extraction network design for the front-end network. We focus on how to design an efficient back-end network. In fact, the module proposed in this work can be plugged into other feature-extraction networks, such as MobileNet~\cite{mobinenetv2} and ShuffleNet~\cite{shufflenet}.

\subsection{General Deep Network Acceleration}
In addition to the real time network for semantic segmentation research, there are extensive studies on general deep network acceleration for applications such as object detection and image classification. For example, network quantization~\cite{Jung_2019_CVPR} is applied for convolution parameters for better inference speed than the floating point computation. On the other hand, network compression technique either uses a pruning operation~\cite{luo2017thinet} to reduce the network structure or use a bigger network to guide the training of a smaller network~\cite{hinton2015distilling}. Recently, many light-weight backbone networks, such as MobileNet~\cite{mobinenetv2} and ShuffleNet~\cite{shufflenet}, are also proposed for efficient feature extraction with light-weight building blocks. Note that these research works are orthogonal to our work. The proposed network can benefit from the advance of these directions.

\section{Approach}
\subsection{Approach Overview}
Given an input RGB image $I\in\mathbb{R}^{H'\times W'\times3}$, deep learning based semantic segmentation method usually consists of successive convolution layers and the output is $L\in\mathbb{R}^{H\times W\times N}$, where $N$ denotes the number of class for the segmentation task, and $L$ is the class probability for each pixel. $H$ and $W$ denotes the height and width of the image, respectively. Differently, in our formulation we allow low-resolution input and high-resolution output to accelerate the process so we have $H'\leqslant  H$ and $W' \leqslant  W$. The network contains the front-end and back-end as shown in Fig.~\ref{fig:idea}. The front-end is for feature extraction, where we can employ existing deep models, such as VGG~\cite{simonyan2015very}, ResNet~\cite{resnet}, and MobileNet~\cite{mobinenetv2}. We can extract the feature from one or more intermediate layers from these networks. Since this part is not the main focus for this paper, we do not explain it in detail. For the back-end network, the input feature map is extracted from the front-end and the output is the probability map $L$. Here we employ the proposed cascaded factorized ASPP (see Sec.~\ref{sec:cfaspp}) that is fused with feature space super-resolution (see Sec.~\ref{sec:sr}).

\subsection{Cascaded Factorized ASPP}
Due to the variations of the object size, how to capture and fuse image feature in different scales is important for semantic segmentation. Atrous convolution~\cite{deeplabv3} has been employed extensively for semantic segmentation. Different from conventional convolution filters, atrous convolution filters can be treated as inserting $r-1$ zeros between two neighboring filter values along each spatial dimension, where $r$ is the atrous rate. We can see that atrous convolution allow us to enlarge the filter's \textit{field-of-view} without increasing the number of model parameters and computation cost. On the other hand, inspired by spatial pyramid pooling method of~\cite{spp}, Chen~\etal~\cite{deeplab} propose atrous spatial pyramid pooling (ASPP) for semantic segmentation. In particular, in ASPP, there are multiple parallel $3 \times 3$ atrous convolutions with different atrous rates. Those parallel atrous convolutions are applied on top of the feature map from a feature extraction network. We can see that the multiple atrous rates can help the network to capture multi-scale context information. However, the ASPP is applied on the feature map obtained from a deep network. The channel number of the feature map is usually large (\eg., 512 for ResNet 18~\cite{resnet}). Even with the kernel size of $3\times 3$, ASPP still consumes substantial computation cost. To reduce this problem and further increase the effectiveness of ASPP, we do the following modifications.

\begin{figure*}[t]
	\centering
	\includegraphics[width=\linewidth]{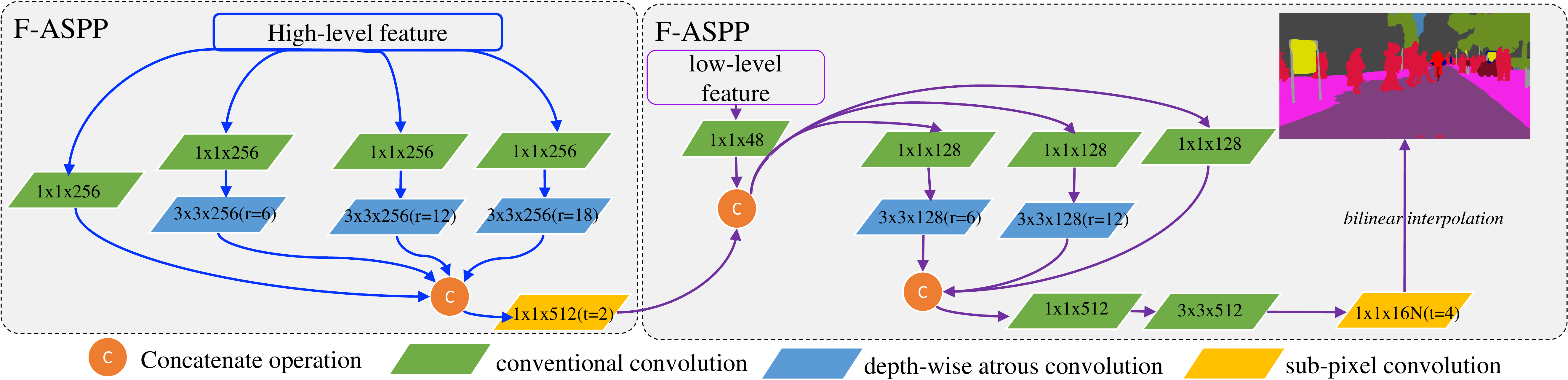}
	\caption{Illustration of the proposed back-end network for semantic segmentation. The high-level and low-level feature is obtained from the deeper and shallower layers of the front-end network, respectively, to capture the image description in different levels. The numbers on the convolution layer denotes the according parameters (\eg., `$3 \times 3 \times 256 (r=18)$' means that the kernel size is $3 \times 3$ with output channel 256. $r$ is the atrous rate and $t$ is the up-sampling factor.) $N$ is the class number of the segmentation task. For example, the last sub-pixel convolution layer with up-sampling factor $t=4$ will generate a feature map with $N$ channels. The spatial size is 4x larger than that of the previous feature map. Then each channel describes the probability of the according class. The class label can be obtained by an `argmax' operator. Each convolution layer is followed by batch normalization~\cite{573696ce6e3b12023e5ce95a} and Relu nonlinear activation~\cite{krizhevsky2012imagenet}.}
	\label{fig:network}
\end{figure*}
Firstly, we decompose the $3\times 3$ atrous convolution layer into two layers: 1) point-wise convolution layer (\ie, kernel size is $1\times 1$) that linearly combines the input channels and reduce the dimension of the output channel dimension; This layer is to perform channel-wise information interaction; 2) depth-wise and atrous convolution with the same kernel size (\ie, $3\times 3$) and atrous rate as the original atrous convolution. The depth-wise convolution here is to reduce the computation cost. This layer is to enable the model to capture the feature of the neighboring area. Similar factorization is also used in~\cite{mobinenetv2,resnet}. Differently, depth-wise convolution is employed here instead of conventional convolution. To this end, we have the factorized ASPP (F-ASPP) module. Let $F_i$ and $F_o$ be the input and output channel number, we can see that for the original $3 \times 3$ atrous convolution, the computation complexity for a feature map $M_{i}\in\mathbb{R}^{h\times w\times F_i}$ is $h \times w \times 3 \times 3 \times F_i \times F_o$, while the computation complexity of the factorized one is $h \times w \times F_i \times F_o$+$h \times w \times 3 \times 3\times F_o$. Given that $F_i$ and $F_o$ is 512, and 256 respectively in our implementation, we can see that the computation cost is about 8.8x reduced. 

Secondly, instead of applying the ASPP module only once, we cascade two factorized ASPPs in our network. The motivation here is that we can perform extensive multi-scale context aggregation in this case. On the other hand, the factorize ASPP already reduce the computation a lot compared to the original ASPP and we use less channels in the second F-ASPP, cascading this component does not bring much additional computation cost but improve the accuracy significantly (see our experiment). The detailed structure of the CF-ASPP is illustrated in Fig.~\ref{fig:network}.

\label{sec:cfaspp}
\subsection{Feature Space Super-resolution}
\label{sec:sr}
In addition to kernel factorization, a simple way to reduce the computation cost is to reduce the input image resolution and up-sample the low-resolution result to the high-resolution one. However, recovering high-resolution result from low-resolution is challenging. Another issue is that after several stages of the front-end network, the feature map spatial size also decreases a lot (\eg., 1/8, 1/16 of the input size). This is mainly because 1) the front-end network contains sub-sampling layers (\eg., max-pooling) or layers with stride lager than 1; 2) the channel number in deep layer is usually large (\eg, 512, 1024)) and keeping high-resolution feature map in deep layers will bring too much computation cost. So we need to generate a high-resolution segmentation map using the low-resolution input. A parameter-free interpolation operation (\eg., bilinear up-sampling) may not be a good solution for this.

In this work, we treat this problem as a super-resolution process in the feature space. Firstly, in the training process, we use down-sampled RGB image as input, and the original high-resolution class label map as the ground truth. Secondly, for the network structure design, we gradually up-sample the feature map in the back-end network. The up-sample operation is performed by sub-pixel convolution~\cite{7780576} that is widely used for the image super-resolution task. Here we explain how we apply sub-pixel convolution in our work. Given the input feature map $M_i\in\mathbb{R}^{h\times w \times F_i}$ and the up-sampling factor $t$, we need to generate an output feature map $M_o\in\mathbb{R}^{tw\times th\times F_i}$. Firstly, we apply a pixel-wise convolution layer to $M_i$ and generate a feature map $M_m\in\mathbb{R}^{h\times w \times F_i t^{2}}$. Secondly, we apply a periodic shuffling operator that rearranges the elements of the feature map $M_m\in\mathbb{R}^{h\times w \times F_i t^{2}}$ to a feature map $M_o\in\mathbb{R}^{tw\times th\times F_i}$. Detailed definition of periodic shuffling can be found in~\cite{7780576}. It has been explained in~\cite{subpixelconv2} that compared to the deconvolution operator with the same computation budget, sub-pixel convolution has more representation power. Detailed explanation for this is out of the scope of this paper. 

To this end, we have the sub-pixel convolution for feature map super-resolution. Here we apply this operation twice in the proposed CF-ASPP. Fig.~\ref{fig:network} illustrates the network details. 

\section{Experiments}
\subsection{Evaluation Dataset}
\begin{figure}[t]
	\centering
	\includegraphics[width=\linewidth]{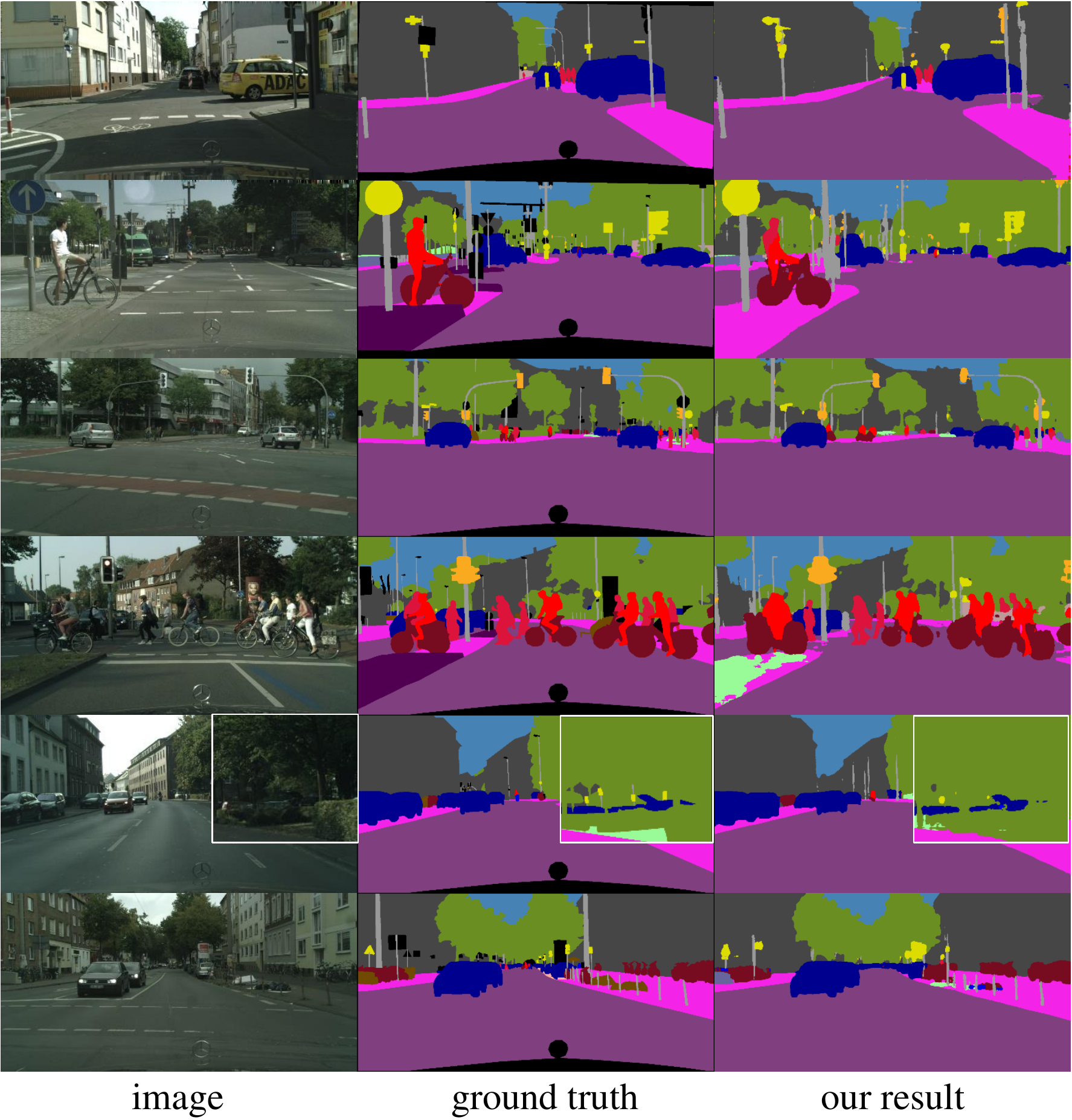}
	\caption{Illustration of our result on the cityscapes validation set.}
	\label{fig:cityscapes}
\end{figure}

We perform the evaluation on the cityscapes dataset~\cite{Cordts2016Cityscapes} since it is a popular and standard benchmark for semantic segmentation research. The high resolution input (i.e., 1024$\times$2048 RGB image) makes it challenging for real time segmentation algorithms. We follow the official evaluation protocol of this benchmark. In particular, it contains an image collection with fine annotations of 30 common classes in urban street scenes (e.g., road, car, sky, person). The collection is divided into training, validation, and test split, with 2,975, 500, and 1,525 images, respectively. Images in this dataset are illustrated in Fig.~\ref{fig:cityscapes}. Among the 30 annotated classes, 19 classes are used for evaluation. According to the dataset, the official accuracy criterion is the mean intersection-over-union (IoU) metric. In particular, we have $IoU = TP/(TP+FP+FN)$, where TP, FP, and FN denotes the size of true positive, false positive, and false negative, respectively. For the inference speed criterion, we use the direct metric--inference time. We do not use FLOPs because it is an indirect metric. It is usually not equivalent to the direct metric because it cannot reflect some hardware related factors such as memory access cost and degree of parallelism. This has been discussed in~\cite{shufflenetv2}.
\subsection{Implementation Details}
For the software platform, the proposed algorithm is implemented by Pytorch 1.1 with CuDNN v7.0 (no TensorRT or other inference optimizers to avoid external influence). For the hardware, the program runs on a PC with Nvidia Titan X (Maxwell) GPU. This is to keep the same platform with many other current methods~\cite{599c7972601a182cd263ec00,fastscnn,contextnet} for fair comparison. When we evaluate the running time performance, we use a single CPU thread and a single GPU and we measure the average model inference time one the 500 cityscapes validation images.

In the training process, for the front-end network, we use ResNet-18~\cite{resnet} pre-trained with the ImageNet classification task~\cite{imagenet}. For the high-level feature from the front-end network, we use the feature from the last convolution layer before the global average pooling layer~\cite{resnet}. For the low-level feature, we use the feature from the layer `conv3\_x' of the network. To train the whole network, we use the 2,975 training images with fine annotations from cityscapes dataset. Batch normalization~\cite{573696ce6e3b12023e5ce95a} and Relu activation~\cite{krizhevsky2012imagenet} is used after every convolution layer. Stochastic gradient decent (SGD) with momentum 0.9 and batch-size 16 is used in the training process. The initial learning rate is set 0.1 and decayed by a factor of 0.9 every 50 epochs. All the experiments of the proposed method are trained for 400 epochs. We also perform extensive data augmentation as other current methods, including random flipping, rotation, color channel noise, and resizing. Similar to other methods, we use the cross-entropy loss.

\begin{table}[t]
\caption{Ablation Study for the Proposed Method on the Cityscapes Validation Set with Nivida Titan X (Maxwell)}
\label{tab:ablation_study}
\begin{center}
\begin{tabular}{c|c|c|c|c}
\hline
Test ID&Method ID&input size&mIoU (\%)&running time (ms)\\
\hline \hline
1& 1& 1024 $\times$ 2048&72.01&42.0\\
2& 4& 1024 $\times$ 2048&73.52&46.1\\
3& 1& 768 $\times$ 1536&70.40&25.4\\
4& 4& 768 $\times$ 1536&72.28&26.8\\
5& 1& 512 $\times$ 1024&68.33&15.8\\
6& 2& 512 $\times$ 1024&68.61&13.3\\
7& 3& 512 $\times$ 1024&70.05&13.5\\
8& 4& 512 $\times$ 1024&70.29&14.6\\
9& 4& 512 $\times$ 768&69.84&11.9\\
\hline
\end{tabular}
\end{center}
\end{table}

\subsection{Ablation Study on Network Structure}
To evaluate the proposed method, firstly we perform ablation study using the following baseline methods:
\begin{enumerate}
\item Front-end network with ResNet-18 and back-end network with original ASPP and the decoder from DeeplabV3+~\cite{deeplabv3p}.
\item Front-end network with ResNet-18. The back-end network contains one F-ASPP (without feature space resolution) and the decoder from DeeplabV3+~\cite{deeplabv3p}.
\item Front-end network with ResNet-18. The back-end network contains CF-ASPP (without feature space resolution) and no decoder from DeeplabV3+~\cite{deeplabv3p}.
\item The full proposed method.
\end{enumerate}

From Tab.~\ref{tab:ablation_study}, we can have the following observations by comparing different test cases. By comparing Test 1, 2, 3 and 4, we can see that the accuracy of proposed back-end network is better than the original one from DeeplabV3+~\cite{deeplabv3p}. In addition, the accuracy decreases when the input resolution decreases. But the accuracy degradation is smaller for the proposed method. This demonstrates the effectiveness of the proposed feature space super-resolution approach. Although our running time consumption is larger than baseline method 1 for high resolution input, we can see that the situation reverses when given lower resolution input (see Test 5 and 8). This is mainly because the additional computation of the proposed method is caused by the increased number of layers. Given high resolution input, the feature map spatial size is large, memory transfer between layers consumes a large part. Given lower resolution input, the situation reverses. This also shows that our method fits the cases with smaller input size.

By comparing Test 5, 6, 7, and 8, we can see that: 1) By incorporating F-ASPP, the performance can be improved a little, and the improvement can be larger if we employ CF-ASPP. This demonstrates the effectiveness of F-ASPP and cascading F-ASPP. 2) F-ASPP is faster than the original ASPP (Test 5 and 6). 3) With CF-ASPP, both the running time efficiency and accuracy is better than the original one (Test 5 and 7). 4) By feature space super-resolution, we can obtain better result (Test 7 and 8).

\begin{table}[t]
\caption{Comparison with Other Methods on Cityscapes Test Set (the running time is measured with Nivida Titan X (Maxwell)\protect\footnotemark[1]}
\label{tab:cityscapes}
\begin{center}
\begin{tabular}{c|c|c}
\hline
Method &mIoU (\%)&running time (fps)\\
\hline \hline
SegNet~\cite{segnet}&56.1&16.7\\
ENet~\cite{enet}&58.3&20.4\\
ICNet~\cite{599c7972601a182cd263ec00}&69.5&30.3\\
ERFNet~\cite{erfnet}&69.7&41.7\\
ContextNet~\cite{contextnet}&66.1&41.9\\
BiSeNet~\cite{bisenet}&68.4&105.8*\\
GUN~\cite{gun}&70.4&33.3*\\
Fast-SCNN~\cite{fastscnn}&68.0&75.3\\
ESPNetV2~\cite{espnetv2}&66.4&-\\
ThunderNet~\cite{thundernet}&64.0&96.2*\\
LEDNet~\cite{lednet}&70.6&71.0**\\
Ours ($512\times1024$ input)&70.2&68.5\\
Ours ($512\times768$ input)&68.4&84.0\\
\hline
\end{tabular}
\end{center}
\end{table}
\footnotetext[1]{For the running time, `*' means that the time is measured using Nvidia Titan Xp and `**' means Nvidia GTX1080Ti. It is benchmarked in~\cite{cnnbenchmark} and~\cite{fastscnn} that these two GPUs are about 60\% faster than Nvidia Titan X (Maxwell) for the network forward time.}

\subsection{Comparison with Other Methods}
We also compare the proposed the method with other state-of-the-art real-time semantic segmentation algorithms including: ENet~\cite{enet}, ICNet~\cite{599c7972601a182cd263ec00}, ContextNet~\cite{contextnet}, ESPNetV2~\cite{espnetv2}, GUN~\cite{gun}, Fast-SCNN~\cite{fastscnn}, ERFNet~\cite{erfnet}, BiSeNet~\cite{bisenet}, LEDNet~\cite{lednet} and ThunderNet~\cite{thundernet}. The results are listed in Tab.~\ref{tab:cityscapes}. The results of other method are from related literatures. We can see that we almost achieve the fastest speed compared to other methods. Although some methods have faster FPS records, but they are evaluated on other GPUs that have higher performance. For example, a same network of~\cite{fastscnn} runs 75.3 fps and 123.5 fps on Titan X (Maxwell) and Titan Xp, respectively~\cite{fastscnn} (\ie., 68\% faster). For the accuracy, our method is also among the leading competitors. This shows that our method achieves the state-of-the-art latency-accuracy trade-off. 

In addition, we test the inference time of the proposed network in a Jetson AGX Xavier embedded system and obtain 21.2fps using the input image size of 512 $\times$ 768. Here we simply test it with an unoptimized implementation (we only use the python interface of Pytorch and do not use the TensorRT library or NVDLA engine for inference acceleration), thus the running time can be further reduced by other engineering adaptation.
\subsection{Qualitative Analysis}
Fig.~\ref{fig:result_fig} shows some of our result on the cityscapes validation set. We can see that the proposed method can deal with some of the small and thin objects. From the 5th row, we can see that even some cars are occluded, they can still be identified to some extent. This figure is best viewed with zoomed in.

\section{Conclusion}
This paper proposes a back-end network component for real time semantic segmentation. On one hand, we propose a cascaded factorized ASPP module for efficient multi-scale context aggregation. On the other hand, to allow the model to use down-sampled lower-resolution image as input and generate high-resolution segmentation output, we employ the super-resolution approach in the feature space. We conduct extensive experiments and the effectiveness of these two aspects have been demonstrated. The latency-accuracy trade-off outperforms many existing state-of-the-art methods. Future research direction can be how to combine the proposed network with other network acceleration techniques such as network compression and quantization.
\bibliographystyle{plain}
\bibliography{egbib}
\end{document}